\documentclass{article}
\usepackage{ijcai20}

\usepackage{times}

\usepackage{soul}
\usepackage{url}
\usepackage[hidelinks]{hyperref}
\usepackage[utf8]{inputenc}
\usepackage[small]{caption}
\usepackage{graphicx}
\usepackage{amsmath}
\usepackage{booktabs}
\usepackage{color}
\usepackage{multirow}

\usepackage{algorithm}
\usepackage{algorithmic}
\title{\textbf{PCA-SRGAN: Incremental Orthogonal Projection Discrimination for Face Super-resolution}}
\author{
Hao Dou$^{1,2}$\and
Chen Chen$^{1,2}$\and
Xiyuan Hu$^{3}$\footnote{Corresponding Author}\and
Zuxing Xuan$^{4}$\and
Zhisen Hu$^{5}$\and
Silong Peng$^{1,2}$\\
\affiliations
$^1$Institude Of Automation, Chinese Academy Of Sciences\\
$^2$The Chinese academy of science\\
$^3$Nanjing University of Science and Technology\\
$^4$Beijing Union University\\
$^5$Beijing University of Posts and Telecommunications\\
\emails
\{douhao2015, chen.chen,silong.peng\}@ia.ac.cn,
xiyuan.hu@foxmail.com,
zuxingxuan@163.com,
huzhisen1117@gmail.com
}

\begin{document}

\maketitle

\begin{abstract}
Generative Adversarial Networks (GAN) have been employed for face super resolution but they bring distorted facial details easily and still have weakness on recovering realistic texture. To further improve the performance of GAN-based models on super-resolving face images,  
we propose PCA-SRGAN which pays attention to the cumulative discrimination in the orthogonal projection space spanned by PCA projection matrix of face data.
By feeding the principal component projections ranging from structure to details into the discriminator, 
the discrimination difficulty will be greatly alleviated and the generator can be enhanced to reconstruct clearer contour and finer texture, helpful to achieve the high perception and low distortion eventually. 
This incremental orthogonal projection discrimination has ensured a precise optimization procedure from coarse to fine and avoids the dependence on the  perceptual regularization. % widely-used (VGG) 
We conduct experiments on CelebA and FFHQ face datasets. The qualitative visual effect and quantitative evaluation have demonstrated the overwhelming performance of our model over related works. 

\end{abstract}

\section{Introduction}
Face image super-resolution (FSR), also known as face hallucination, aims to recover a high-resolution (HR) face image from its low-resolution (LR) input, which is helpful to numerous facial analysis applications. %,\cite{fsrnet}
For many years image super-resolution (SR) has been a challenging research problem and data-driven deep learning draws a lot of attention recently. Deep learning has been employed to achieve an end-to-end single image super-resolution model, such as the Convolutional Neural Networks (CNN) based methods (e.g. SRCNN \cite{SRCNN}, SRResnet \cite{SRResnet} ) and recent approaches using Generative Adversarial Networks (GAN) \cite{GAN} (e.g. SRGAN \cite{SRGAN}, ESRGAN \cite{ESRGAN}), which out-perform most of traditional methods. %\cite{Benchmark}. 
Specifically for face image super resolution task, many types of face priors are combined and applied to deep-learning-based SR methods. Most of them learn a straight mapping by a variety of regularizations, including facial masks \cite{fsrnet}, attributes \cite{attributes}, etc.

\begin{figure}[tp]
	%\centering
	\centerline{
		\includegraphics[width=0.49\textwidth,height=0.21\textwidth]{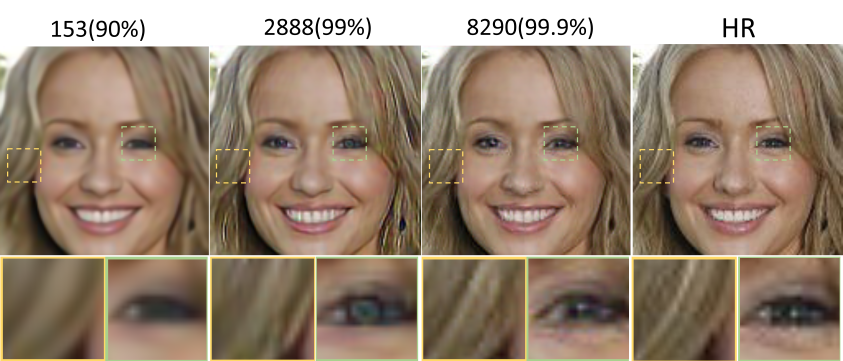}} %
	\caption{The reconstruction results of our PCA-SRGAN by incremental discrimination along with the gradually increased subspace. Above the images list the corresponding subspace dimensions and energy proportions.
	} %
	\label{Fig 1} %
\end{figure}

Compared to CNN-based models which aim at the less distortion, GAN-based SR models have demonstrated the superior perceptual performance. Extra perceptual loss like VGG \cite{perceptual,SRGAN} are often accompanied to regularize the generator for more stable results with less artifacts.
However, GAN-based models still have weakness on generating realistic and precise texture and produce fake details easily especially for face images which lead to the distortion.
To enhance the performance of GAN-based models, cumulative learning is considered as a effective strategy and introduced to super resolution task to lower the reconstruction difficulty.
Typical cumulative learning for super resolution concentrates on the incrementally network structure -- progressively grows both the networks and the upscale factors \cite{ProSR} or pyramid levels \cite{LAPSRN}.
Different from this kind of cumulative training in which the whole networks grow for different stages, we can maintain the network structure and seek a new cumulative training method for adversarial learning --- incrementally
discriminate the different levels of information or components for easier and more precise reconstruction.
 
As one of the classical statistical methods, principal component analysis (PCA) \cite{PCA} can give a perfect hierarchical representation of face images, namely the orthogonal projections on the PCA orthogonal subspace.
%which is very suitable for face image processing. 
%An orthogonal space can be determined by the PCA projection matrix. 
We can train the GAN-based SR model by discriminating the projections of face images on the PCA subspace cumulatively to reconstruct the super-resoluted images .
Therefore we propose a novel cumulative discrimination and reconstruction approach in this paper referred as \textit{PCA-SRGAN} and focus on the \textit{Incremental Orthogonal Projection Discrimination} in the PCA orthogonal subspace. % projectionfor face super resolution task. 
The orthogonal projections are obtained by projecting both generated SR images and HR images into the PCA subspace.  The corresponding orthogonal projections containing hierarchical facial components from structure to details will be fed into the discriminator to guide the precise optimization and fine-grained reconstruction of generator. 
In this way, our proposed model can lighten the difficulty of discrimination, stabilize the procedure of adversarial training  and enhance the performance of generator, which ensures the ability to produce perceptual and realistic texture while brings relatively low level of distortion and artifacts without any help of auxiliary perceptual regularization.
%will handle the difficulty of trade-off between perception and distortion -- 
An example of cumulative training process has been shown in Figure 1. With the growth of subspace proportion, our model has generated the super-resolved face with more realistic details gradually.

In summary, the main contributions of this work are presented as follows:
%we develop a incremental discrimination and reconstruction approach in PCA orthogonal projection subspace for face super resolution with the help of PCA statistics prior. For brevity, we refer to our method as PCA-SRGAN.without any other feature loss.
\begin{itemize}
	\item we propose a novel cumulative learning strategy for the GAN-based SR method utilizing the incremental orthogonal projection discrimination in the PCA subspace to enhance the face SR task. %which can make full use of the PCA statistics prior.
	\item Our model provides a precise and stable adversarial training method without the need for auxiliary perceptual (VGG) loss, which achieves compelling visual effect with better perception-distortion trade-off than other methods.
	
\end{itemize}

%\iffalse
\begin{figure*}[tp]
	%\centering
	\centerline{
		\includegraphics[width=0.80\textwidth,height=0.175\textwidth]{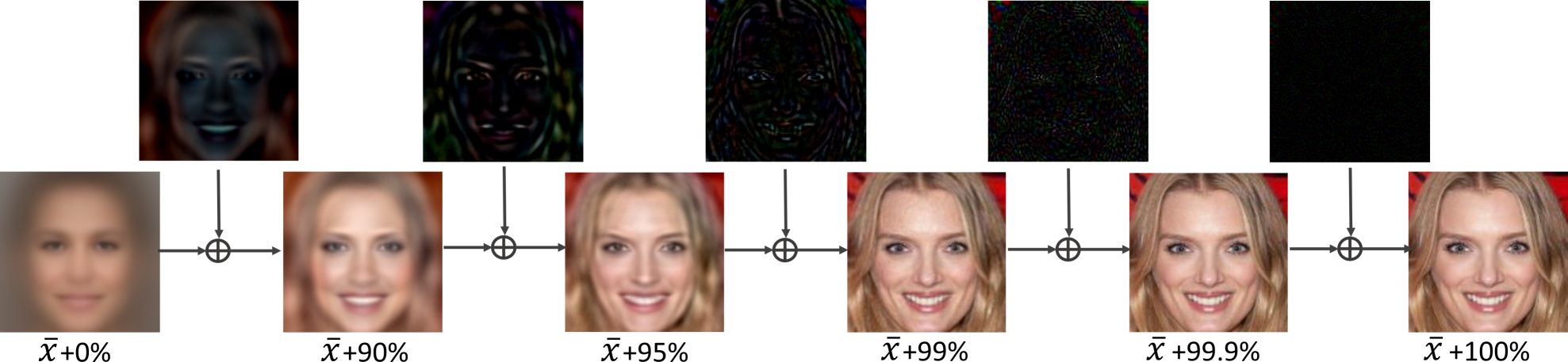}} %
	\caption{
			Visualization for different proportions of principal component projections which have been added to the average face $\overline{x}$. With the growth of proportion, the projections will contain increasing facial information from coarse to fine.  %The residue have been enhanced for easier review or visualization.% equally.
	} %
	\label{Fig 1} %
\end{figure*}
%\fi

\section{Related Works}

\subsection{Face super resolution}
%Single image super-resolution (SISR) is a task to restore the original high-resolution (HR) image from a single low-resolution (LR) image. 
Face super-resolution,
%a.k.a. face hallucination,
is a special case of single image super-resolution (SISR).
The SISR problem has been extensively studied in the literature using a variety of deep-learning-based models. SRCNN, EDSR \cite{EDSR} and SRResnet employ deep convolutional network with various structures for super-resolution. Many works, e.g. SRGAN, ESRGAN, EPSR \cite{EPSR}, and RankSRGAN \cite{RankSRGAN}, also introduce the generative adversarial network to produce perception-oriented results. 
The challenge PIRM2018 \cite{PIRM} has been also held to evaluate the SR methods by jointly
quantifying the accuracy and perceptual quality, namely the trade-off between distortion and perception.  
Based on commonly used SISR methods above, the facial priors are often exploited and combined for face super resolution.
There are many types of prior such as masks, landmarks, heatmaps, applied to the face super resolution \cite{mask,component}. %\cite{heatmap}
Yang decomposes face images into facial components which can be super-resolved separately and fused to the complete face. Adrian Bulat learns to localize the facial landmarks as a constraint for face super resolution \cite{landmarks}. Yu Chen employs an extra network to predict the prior such like facial masks to enhance the performance for very low resolution face \cite{fsrnet}. Besides the structural prior mentioned above, statistical information has been explored for face hallucination. Principal components analysis is a classical statistical prior especially useful for human face problem. Wang replaces the base vectors of PCA dictionary for LR images with that for HR images and learn the coefficients mapping between LR and HR \cite{Eigentrans}. In the view of multi-scale space-frequency domain, Huang designs a wavelet-based network to reconstruct the facial high-frequency information \cite{WaveletSR}. 
Among all kinds of face prior, PCA contains all the necessary information to build a real face and can be applied on end-to-end networks easily.

\subsection{Cumulative learning}
Cumulative learning is a effective learning strategy for complex learning task which refers to starting from an easier subtask and gradually increasing the task difficulty. It is very suitable for the image generation task, including the image super-resolution.
ProGAN \cite{ProGAN} enlarges the network and simultaneously increases the resolution of generated images for easier convergence and better image generation for GAN models. Similarly, Wang performs SR starting from $2\times$ upsampling and then blending with the portions of $4\times$ or larger scaling factors \cite{ProSR}, which are progressive both on architectures and training procedure. Deokyun Kim \cite{ProFSR} introduces attention mechanism to the progressive face SR for very small face images. %PGGAN \cite{PGGAN} and 		
Different from the progressive method by increasing the up-sampling scale, %there are other types of cumulative learning for super resolution.
LapGAN \cite{LAPGAN} and LapSRN \cite{LAPSRN} decompose the generation process by the means of the Laplacian pyramid and train the generator progressively at every level of the pyramid.  %LapGAN \cite{LAPGAN} and 
%PPON\cite{PPON}employ two learning stage of architectures, the first of which aims at high PSNR performance and on this basis another is mounted to pursue good perception.
Compared to common training procedure, the curriculum learning strategy can greatly reduce the training difficulty.
Inspired by this, we propose a new type of cumulative learning -- incrementally learn the discrimination for the PCA projection of face images.

\section{Proposed Method}
GAN-based models have weak performance on super-resolving realistic and perceptual face images. 
To address the issue, we lower the learning difficulty of GAN by incrementally discriminating the SR image's projection on the subspace spanned by the elements of the PCA dictionary generated for HR images.
%Given the PCA dictionary generated for HR images, we will incrementally discriminate and reconstruct the SR image's projection on the subspace spanned by the elements of the dictionary. %and incrementally discriminate and reconstruct the corresponding orthogonal projection.
In this section, we firstly introduce the properties of orthogonal projection on principal component space, then describe the model structure and loss functions. At last, the overall algorithm will be summarized.
\subsection{PCA orthogonal projection}
Principal component analysis (PCA) is a statistical procedure using an orthogonal transformation to convert a set of observations of possibly correlated variables into a set of values of linearly uncorrelated variables called principal components. 

Given a set of HR face samples $\left \{ x_{1}^{h},x_{2}^{h},x_{3}^{h},...,x_{m}^{h}\right \}$, among them $x_{i}^{h}$ is a image vector which contain $n$ pixels and $m$ is the number of samples.
Firstly, 

all samples are centralized by zero-mean normalization.
\setlength\abovedisplayskip{0.002\textwidth}
\setlength\belowdisplayskip{0.002\textwidth}
\begin{align}
\begin{split}
X = \left [ x_{1}^{h} - \overline{x},x_{2}^{h} - \overline{x},...,x_{i}^{h} - \overline{x} \right ], % \overline{x}
\end{split}
\end{align}%
where $\overline{x}$ denotes the average face. 
We construct the covariance matrix and perform the eigen decomposition.
\setlength\abovedisplayskip{0.002\textwidth}
\setlength\belowdisplayskip{0.002\textwidth}
\begin{align}
\begin{split}
XX^{T}P=\lambda P.
\end{split}
\end{align}
The projection matrix can be obtained from the eigenvectors of the covariance matrix.
\setlength\abovedisplayskip{0.001\textwidth}
\setlength\belowdisplayskip{0.001\textwidth}
\begin{align}
\begin{split}
P = (p_{1},p_{2},...,p_{d}),
\end{split}
\end{align}
where $p_{i}$ is the $i$th eigen vector and we use $\lambda_{i}$ to denote the corresponding eigen value. 
 
A sample can be projected into the orthogonal PCA subspace as the orthogonal projection. 
The dimensions of orthogonal subspace are determined by the vectors' number of the projection matrix.
As shown in Figure $2$, the principal components, or described as the orthogonal projections of a face image on different dimensions of orthogonal subspace, has included different levels of facial information. %, from contour to details.
It will be reasonable to feed these projections which represent face from structure to details into the discriminator to improve the reconstruction of super-resolved faces.

\subsection{Incremental orthogonal projection discrimination}

\begin{figure*}[t]
	%\centering
	\centerline{
		\includegraphics[width=0.88\textwidth,height=0.45\textwidth]{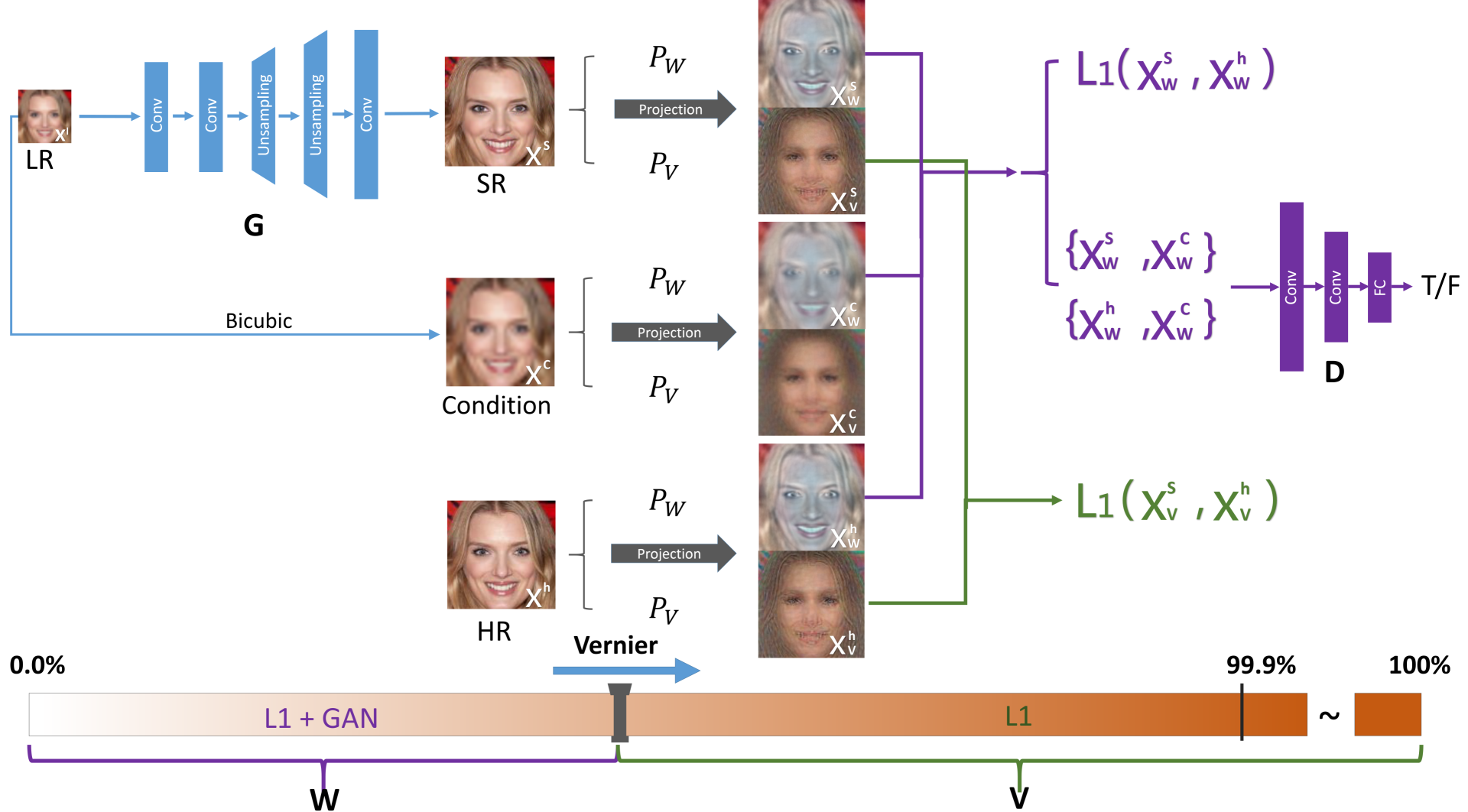}} %
	\caption{The pipeline of the PCA-SRGAN. The progress bar reveals the changing process of subspaces, above which other part illustrates the discrimination or constraint for orthogonal projections on the subspaces.
	} %
	\label{Fig 1} %
\end{figure*}

The pipeline is illustrated in Figure $3$. We perform PCA on HR datasets and get the orthogonal space spanned by the PCA projection matrix $P$. The orthogonal space can be split into two orthogonal complementary subspace $W$ and $V$ respectively determined by $P_{W}$ and $P_{V}$. $P_{W}$ is a projection matrix which contains the first $n$ vectors of $P$ and $P_{V}$ contains the rest vectors. $G$ and $D$ denote the generator and discriminator separately.

Given a generated SR image $x^{s}$ which has been normalized by subtract the mean, $x_{W}^{s}=P_{W}x^{s}$ is the orthogonal projection of $x^{s}$ on the subspace $W$, %=G(x^{l}) 
$x_{V}^{s}=P_{V}x^{s}$ is the orthogonal projection of $x^{s}$ on the subspace $V$. Similar operations are also applied to the HR groundtruth $x^{h}$.
% $x_{W}^{h}=P_{W}x^{h}$ is the orthogonal projection on the subspace $W$, $x_{V}^{h}=P_{V}x^{h}$ is the orthogonal projection on the subspace $V$.
$x_{W}^{s}$ will be sent to the discriminator to guide the adversarial learning for accurate face SR reconstruction. Both of the $x_{W}^{s}$ and $x_{V}^{s}$ will be constrained by the L1 loss. %point-wise
The LR input will be bicubic-upsampled as the condition to assist the discrimination of GAN.
We also illustrate the incremental training procedure by the progress bar shown in the lower part of Figure $3$. %
The vernier will slide from the left to the right side during the training procedure, which means the subspace $W$ will expand from zero to the almost full space, while the opposite situation for subspace $V$. Corresponding to the visualized projections in Figure $2$, the growing subspace $W$ will absorb more and more detailed and precise face components prepared for discrimination, helpful to reduce the learning difficulty of discriminator and make the generator catch the distribution of original HR data progressively. 
Meanwhile, the shrinking subspace $V$ is regarded as a repository which stores the constrained residual information to be added into $W$. %complementary and % and discriminated.  
This curriculum learning strategy guarantees a stable and precise learning process and has the ability to synthesize compelling facial details without visible distortion.
To notice that, the subspace $W$ is increased to $99.9\%$ of the full space and the last $0.1\%$  percentage could be treated as the random noise space and not necessary to be regularized by the GAN model. %no need

\noindent
{\bfseries{Networks structure}}\quad 
The proposed model consists of only two networks, the generator and the discriminator, which construct the elementary structure of GAN. 
%We use the RRDBnet as generator and VGG-like classifier as discriminator, the same to the baseline ESRGAN. 
The same to the baseline ESRGAN, RRDBnet is used as the generator which combines multilevel residual network and dense connections without BN layers, and the discriminator contains eight convolutional layers and two linear layers to generate the classification value.

\subsection{Loss functions}
We use pixel-wised L1 loss and GAN loss without any auxiliary feature-based or perceptual loss to obtain the final results. The GAN loss is applied to the orthogonal projection on the subspace $W$.

\noindent %auxiliary
{\bfseries{Conditional GAN loss}}\quad 
The LSGAN \cite{LSGAN} is applied as basic GAN loss because it brings smooth gradients which is more suitable for incremental training than vanilla GAN. 
Based on the relativistic discrimination used by the ESRGAN, we further lower the difficulty of discrimination for the orthogonal projections by introducing the conditional LR input. The input LR images will be bicubic-upsampled and concatenated with HR or SR images which construct $\left \{x_{W}^{h},x_{W}^{c} \right\}$ and $\left\{x_{W}^{s},x_{W}^{c} \right\}$, and sent to the discriminator. 

We formulate the loss of generator as:
\setlength\abovedisplayskip{0.002\textwidth}
\setlength\belowdisplayskip{0.002\textwidth}
\begin{align}
\begin{split}
L_{G}=&E_{x_{W}^{s}}[(D(\left\{x_{W}^{s},x_{W}^{c}\right\})- E_{x_{W}^{h}}[ D(\left \{  x_{W}^{h},x_{W}^{c} \right\})] -1)^{2}] + \\ &E_{x_{W}^{h}}[(D(\left \{x_{W}^{h},x_{W}^{c}\right\})- E_{x_{W}^{s}}[ D(\left \{ x_{W}^{s},x_{W}^{c}\right\})])^{2} ],
\end{split}
\end{align}
and the loss of discriminator is the symmetrical form :

\setlength\abovedisplayskip{0.002\textwidth}
\setlength\belowdisplayskip{0.002\textwidth}
\begin{align}
\begin{split}
L_{D}=&E_{x_{W}^{h}}[(D(\left\{x_{W}^{h},x_{W}^{c}\right\}) - E_{x_{W}^{s}}[ D(\left \{  x_{W}^{s},x_{W}^{c} \right\})] -1)^{2}] + \\ &E_{x_{W}^{s}}[(D(\left \{x_{W}^{s},x_{W}^{c}\right\})- E_{x_{W}^{h}}[ D(\left \{ x_{W}^{h},x_{W}^{c}\right\})])^{2} ].
\end{split}
\end{align}

The final GAN loss is:
\setlength\abovedisplayskip{0.002\textwidth}
\setlength\belowdisplayskip{0.002\textwidth}
\begin{align}
\begin{split}
L_{GAN}(G,D)=L_{G} + L_{D}.
\end{split}
\end{align}
\noindent
{\bfseries{pixel-constrained L1 loss}}\quad 
We apply the 1-norm distance for the orthogonal projections on two subspaces between SR images and HR images. $\alpha$ implies the different strength of constraint placed on the two orthogonal projections.
\setlength\abovedisplayskip{0.005\textwidth}
\setlength\belowdisplayskip{0.005\textwidth}
\begin{align}
\begin{split}
L_{1}(G) =E_{x_{W}^{s}}[\left \| x_{W}^{s}-x_{W}^{h}  \right \|_{1}]+\alpha  E_{x_{V}^{s}}[\left \| x_{V}^{s}-x_{V}^{h}  \right \|_{1}].
\end{split}
\end{align}

\noindent
{\bfseries{Total loss}}\quad 
We conclude the total loss as 
\setlength\abovedisplayskip{0.005\textwidth}
\setlength\belowdisplayskip{0.005\textwidth}
\begin{align}
\begin{split}
L_{Total}(G,D) = L_{1}(G) + \beta L_{GAN}(G,D),
\end{split}
\end{align}
where the weight $\beta$ is used to balance different loss terms.

\subsection{Algorithm summary}
Let $\phi $ denote the training set which contains $m$ samples of LR face images $\left \{x_{1}^{l},x_{2}^{l},...,x_{m}^{l}\right \}$ and HR face images $\left \{x_{1}^{h},x_{2}^{h},...,x_{m}^{h}\right \}$. 
$\bar{x}$ denotes the average face  of all HR face images in the training set. 
$d$ denotes the dimension of the full space while ${d}'$ denotes the dimension of subspace $W$ when it increase to $99.9\%$ of the full space. The whole algorithm has been summarized in $Algorithm 1$.
\begin{algorithm}[!h]
	\caption{Incremental orthogonal projection discrimination.} 
	\label{alg:Framwork} 
	\begin{algorithmic}[1] %这个1 表示每一行都显示数字
		\REQUIRE ~~\\ %算法的输入参数：Input
		$\phi $, $P$, $d$, ${d}'$, $\bar{x}$, $G$, $D$; The step size $k$ for each epoch.\\ % , $d$ %of subspace dimension growth 
%		Epoch number $e$ on each step.
%		The generator $G$ and the discriminator $D$;\\
%		The training set which contains $m$ samples of LR face images $\left
		
%		\{x_{1}^{l},x_{2}^{l},...,x_{m}^{l}\right \}$ and HR face images $\left
		
%		\{x_{1}^{h},x_{2}^{h},...,x_{m}^{h}\right \}$;\\
%		The PCA projection matrix $P=\left \{p_{1},p_{2},...,p_{d}\right \}$;\\
		%The orthogonal space $W$ determined by $P$;\\
		%The total dimensions $d$ of space $W$;\\
		%orthogonal projection of $y$ on $W$, $Pw*y$;\\
%		The mean face $\bar{x}$ of traning set;\\
%		The incremental step length $k$;\\
%		The generator $G$ and the discriminator $D$;\\
		\ENSURE ~~\\ %算法的输出：Output
		$G$.\\
		\STATE Initialize the dimension of subspace $W$: $n = 0$.\\
		\WHILE{$n \leq {d}'$} 
		\STATE  Start the new epoch: $n=n+k$. %Allocate
		\STATE  Separate the principal component vectors to construct orthogonal complementary subspace $W$ and $V$: %according the sequence of principal components;
		
		 	   \quad $P_{W}=\left \{p_{1},p_{2},...,p_{n} \right \}$; 
	 	   
		 	   \quad $P_{V}=\left \{p_{n+1},p_{n+2},...,p_{d} \right \}$.
		\FOR{each batch of samples $\left \{x^{l},x^{h}\right \}$ in $\phi $} 
		%\FOR{each epoch on current step} 
		%\label{code:fram:}
		%\STATE  Sample each  of samples  $\left \{x^{l},x^{h}\right \}$ in $\phi $.
		\STATE Generate the super-resolved face images: 
		
			    \quad $x^{s}=G(x^{l})$.
		
		\STATE Take the bicubic-upsampled $x^{l}$ as condition $x^{c}$.
			   
%			    \quad $x^{c}=Bicubic(x^{l})$
			   		
		\STATE Subtract the average face:
		
			   \quad $x^{s}\leftarrow x^{s}-\bar{x}$, \quad $x^{c}\leftarrow x^{c}-\bar{x}$, \quad $x^{h}\leftarrow x^{h}-\bar{x}$.
			   			   
		\STATE Get the orthogonal projections of $x^{s}$, $x^{a}$ and $x^{h}$ on the subspace $W$:
		
			   \quad $x_{W}^{s}=P_{W}x^{s}$, $x_{W}^{c}=P_{W}x^{c}$, $x_{W}^{h}=P_{W}x^{h}$,
%		       \quad $x_{W}^{s}=P_{W}x^{s}=\sum_{i}^{n}\left \langle x^{s} , p_{i} \right \rangle p_{i}$,		
%			   \quad $x_{W}^{c}=P_{W}x^{c}=\sum_{i}^{n}\left \langle x^{c} , p_{i} \right \rangle p_{i}$, 		
%			   \quad $x_{W}^{h}=P_{W}x^{h}=\sum_{i}^{n}\left \langle x^{h} , p_{i} \right \rangle p_{i}$,
		
		%\label{code:fram:}
%		\STATE Get the residue, namely the orthogonal projection of on complementary subspace $V$;
		\STATE Get the residues on complementary subspace $V$;

		       \quad $x_{V}^{s}=P_{V}x^{s}$, $x_{V}^{c}=P_{V}x^{c}$, $x_{V}^{h}=P_{V}x^{h}$,	      	
%			   \quad $x_{V}^{s}=x^{s} - P_{W}x^{s}=P_{V}x^{s}=\sum_{j}^{d-n}\left \langle x^{s} , p_{j} \right \rangle p_{i}$
%			   \quad $x_{V}^{c}=x^{c} - P_{W}x^{c}=P_{V}x^{c}=\sum_{j}^{d-n}\left \langle x^{c} , p_{j} \right \rangle p_{i}$ 
%			   \quad $x_{V}^{h}=x^{h} - P_{W}x^{h}=P_{V}x^{h}=\sum_{j}^{d-n}\left \langle x^{h} , p_{j} \right \rangle p_{i}$		
%		\STATE Normalize the scale of $x_{W}^{s}$ $x_{W}^{c}$ and $x_{W}^{h}$:
		
		\STATE Apply loss according to Eq $8$.
		\STATE Back-propagation and update $G$ and $D$.%with the supervision of groundtruth  and update $G$ and $D$; 
			   
%		\STATE Back-propagation the gradients and update the generator and discriminator;
		\ENDFOR
		\ENDWHILE
		\RETURN $G$;
	\end{algorithmic}
\end{algorithm}
\begin{figure*}[!ht]
	%\centering
	\centerline{
		\includegraphics[width=1.0\textwidth,height=0.87\textwidth]{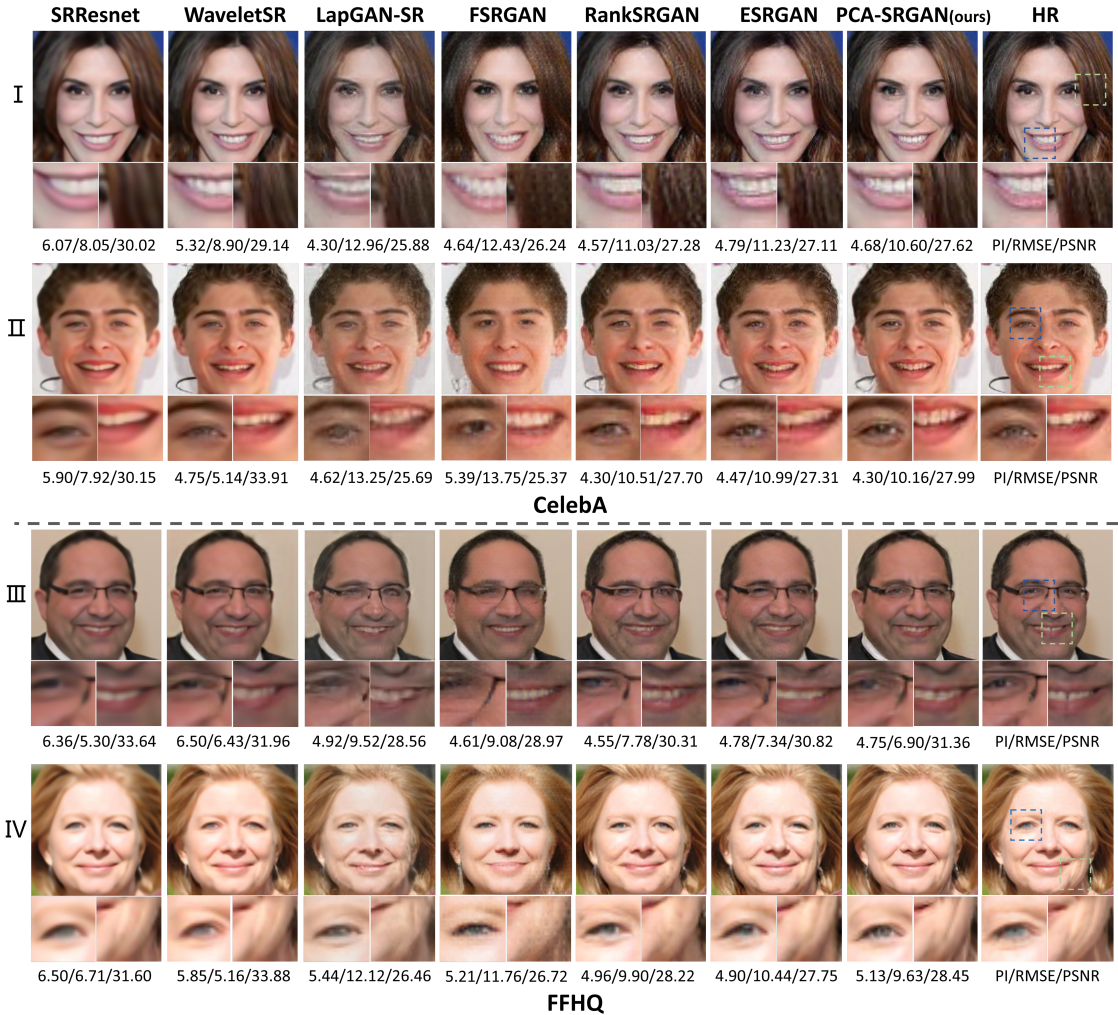}} %
	\caption{Comparisons of related methods on two datasets. %We display several representative examples
	} %
	\label{Fig 1} %
\end{figure*}	

\begin{figure*}[!htp]
	%\centering
	\centerline{
		\includegraphics[width=0.92\textwidth,height=0.29\textwidth]{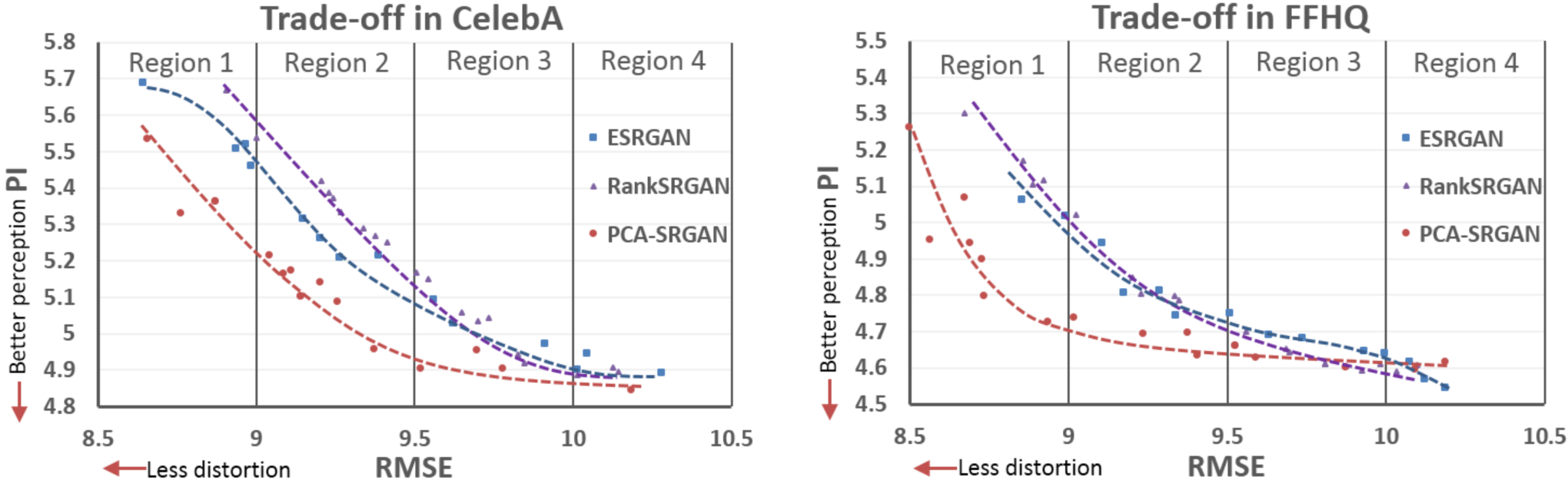}} %
	\caption{
		The trade-off boundaries between perception and distortion for ESRGAN, RankSRGAN and our PCA-SRGAN on two datasets.
	} %
	\label{Fig 1} %
\end{figure*}

\begin{table*}[]	
	\scriptsize	
	\centering		
	\renewcommand\tabcolsep{4pt}
	\label{Tab03}	
	\begin{tabular}{ccccccccccccccc}		
		\toprule		
		\multirow{3}{*}{Method} & \multicolumn{4}{c}{CelebA} & \multicolumn{4}{c}{FFHQ} \\		
		\cmidrule(r){2-5} \cmidrule(r){6-9}
		            & Region 1  &  Region 2		& Region 3   & Region 4 & Region 1  &  Region 2		& Region 3   & Region 4 \\  		
		\midrule		
		ESRGAN &  \textcolor{blue}{5.46/8.98/29.95}   &  \textcolor{blue}{5.21/9.26/29.65}  &  4.97/9.91/29.18    &4.90/10.01/29.07    & \textcolor{blue}{5.02/8.99/29.69}     &\textcolor{blue}{4.74/9.34/29.31}     & 4.64/9.99/28.70   &\textcolor{red}{4.54/10.19/28.59}\\			
		RankSRGAN   & 5.54/9.00/29.86  & 5.25/9.41/29.49  & \textcolor{blue}{4.92/9.85/29.11}  & \textcolor{blue}{4.89/10.01/28.90}   & 5.10/8.89/29.63   & 4.79/9.35/29.28  & \textcolor{red}{4.59/9.93/28.75}   &  \textcolor{blue}{4.59/10.03/28.65} \\						
		PCA-SRGAN   &\textcolor{red}{5.33/8.76/29.96}  &  \textcolor{red}{4.96/9.37/29.38}  &\textcolor{red}{4.90/9.52/29.04}  & \textcolor{red}{4.84/10.19/28.71}   & \textcolor{red}{4.73/8.93/29.66}  &\textcolor{red}{4.63/9.40/29.34}  & \textcolor{blue}{4.60/9.87/28.77}  & 4.60/10.10/28.60  \\		
		\midrule
		Groundtruth & \multicolumn{4}{c}{4.88/0/-} & \multicolumn{4}{c}{4.68/0/-} \\
		\bottomrule		
	\end{tabular}	
	\caption{The values of PI-RMSE and additional PSNR for compared works are provided in preset regions.  We highlight the best performance with red and the second with blue.}
\end{table*}

\section{Experiments}

\subsection{Datasets and preprocess}
We have conducted the experiments on two face datasets.

\noindent
{\bfseries{CelebA}}\quad CelebA consists of a large amount of celebrity face images cropped from the websites \cite{CelebA}. Face images on CelebA are close to images taken in the real scene and contain rich texture, which brings realistic perceptual effect. 

\noindent
{\bfseries{FFHQ}}\quad FFHQ is a high-quality human face image dataset \cite{FFHQ} released recently and contains considerable variation in terms of age, ethnicity and background. Compared to CelebA, face images in this dataset is more clear and smooth with less noise.

Similar to FSRGAN, we select 18000 cropped and aligned face images with the size of $128\times$ 128 as training sets and additional 100 images are used as validation sets for both CelebA and FFHQ.
PCA operations are applied to the training sets of these two datasets. We get the mean and projection matrix for orthogonal projection discrimination on corresponding subspace.

\subsection{Implemtation details}
Following SRGAN and ESRGAN, all experiments are performed with a scaling factor of $4\times$ between LR and HR images. The PSNR-oriented models are first obtained by pre-training on the corresponding face datasets. Then we employ the pre-trained models as an initialization for the generator and the Adam \cite{Adam} as the optimizer. %\cite{adam}

The training process contains 200 epochs divided into two stages according to the dimension of subspace $W$. % or energy of principal components.
In the first 100 epochs, the subspace grow up to $99\%$ of the full space by increasing the dimension evenly and we set the learning rate as 0.0002 with loss weights $\alpha =1.0$ and $\beta =0.02$. % ($lr$)
In next 100 epochs, the subspace grow up from $99\%$ to $99.9\%$ along with a linear decay for learning rate from 0.0002 to 0.00001, and the weights are set as $\alpha =0.0$ and $\beta =0.05$ to enlarge the effect of GAN loss and avoid the over-smooth of L1 loss. 
%Meanwhile a linear decay is also applied for learning rate from 0.0002 to 0.00001 in the second stage. 
With this weight setting of loss functions, our model gets excellent visual effect as shown in Figure 4. Furthermore, the trade-off between perception and distortion of our model can be explored by applying a series of different weight settings.

\subsection{Evaluation}

We evaluate our PCA-SRGAN by qualitative visual effects and quantitative performance of the trade-off between perception and distortion. The PI and RMSE used in the PIRM2018 Challenge are chosen as the quantitative indices for perception and distortion respectively. % trade-off between
PI is a no-reference image quality measure and a lower value indicates better perceptual performance. RMSE is similar to the PSNR and positively correlated with the distortion degree.

\noindent
{\bfseries{Qualitative results}}\quad
We make comparisons with related methods, among which FSRGAN, ESRGAN and RankSRGAN have the aid of the perceptual (VGG) loss. 
As we can see in Figure 4, examples are displayed for visual comparison and the values of PI and MSE are provided as references. Considering the visual effect of whole faces, our PCA-SRGAN has 
%obtained the best performance compared to other methods
outperformed previous approaches, reflected in more natural perception, clear texture and realistic facial components. 
For instance, PCA-SRGAN has produced visually pleasing sharper contour and facial components in all displayed examples than SRResnet which outputs the blurry eyes, teeth or hair and WaveletSR which generates some texture but still results in the smooth and unclear edge.
% get rid of \textbf{compared to SRResnet}
PCA-SRGAN obtains less distortion and generates desired texture which is closer to groundtruth than other GAN-based methods, including those perceptual loss aided models.
LapGAN-SR brings unnatural texture containing unpleasing noise (see Face \uppercase\expandafter{\romannumeral2} and \uppercase\expandafter{\romannumeral3}) which enlarges the distortion. %tends to  and superfluous 
FSRGAN produces warped face obviously in all face images which is likely to be caused by inaccurate facial prior prediction. 
ESRGAN and RankSRGAN achieve good perceptual results but our model obtains better visual effect on local patches, illustrated by clearer shape of the teeth (see Face \uppercase\expandafter{\romannumeral1},
\uppercase\expandafter{\romannumeral2}, and 
\uppercase\expandafter{\romannumeral3}), more lifelike details in the eyes
(see Face \uppercase\expandafter{\romannumeral2} and \uppercase\expandafter{\romannumeral4})
as well as the vivid texture of hair (see Face \uppercase\expandafter{\romannumeral1}) and cheek (see Face \uppercase\expandafter{\romannumeral4}).
%close to the groundtruth,  
The superior visual performance has also approximately agreed with the referenced values below the images.% and referenced values below.

\noindent %  of PI-RMSE
{\bfseries{Quantitative performance}}\quad
Among all contrasted methods,
we mainly seek and compare the boundaries of the PI-RMSE trade-off for three models ---  RankerGAN, ESRGAN and our PCA-SRGAN, which have shown obviously better visual effect in Figure 4 than other models.
We choose a number of network weights that yield the PI values on the border over a certain range of RMSE which has been divided into four regions following the rules of the PIRM2018 Challenge. 
The PI and RMSE values construct the plotted points and we draw a trend line along the boundaries in the Figure 5. For each model the best PI value in each region is also selected in the Table 1 for comparison.
%In each region there list the best perception values on the Table 1.
In general, our model has achieved the superior performance of trade-off than ESRGAN and RankerGAN. 
For CelebA dataset, PCA-SRGAN overwhelms them by a much better trade-off boundary on whole regions.
For FFHQ dataset, PCA-SRGAN has a comparable performance --- ranking 1st on the first two regions and ranking 2rd on Region 3.
% and 4 while better on the first two regions.  % with ESRGAN
It's also reasonable for our model to have slight weakness in Region 3 and 4 for the second dataset. 
%The HR images in FFHQ are relatively smooth while our model has caught enough components of them and 
Referring to the PI values of the groundtruth in Table 1, our model which aims to catch enough components of HR face has a convergence of perceptual index and would not produce the PI values far lower than the groundtruth boundlessly. %, different from other methods.
%and produced less fake texture than other methods when the GAN weight is too large, resulting in the little drop of PI index in Region 4. 

%\subsection{Ablation study}
% loss 和 模型是一个整体\textsl{}，没有必要拆分开实验

\section{Ablation study}
We have conducted experiments on the CelebA dataset as the ablation study. All of them share the same network structure, learning rate and training process.

\begin{figure}[!htp]
	%\centering
	\centerline{
		\includegraphics[width=0.50\textwidth,height=0.41\textwidth]{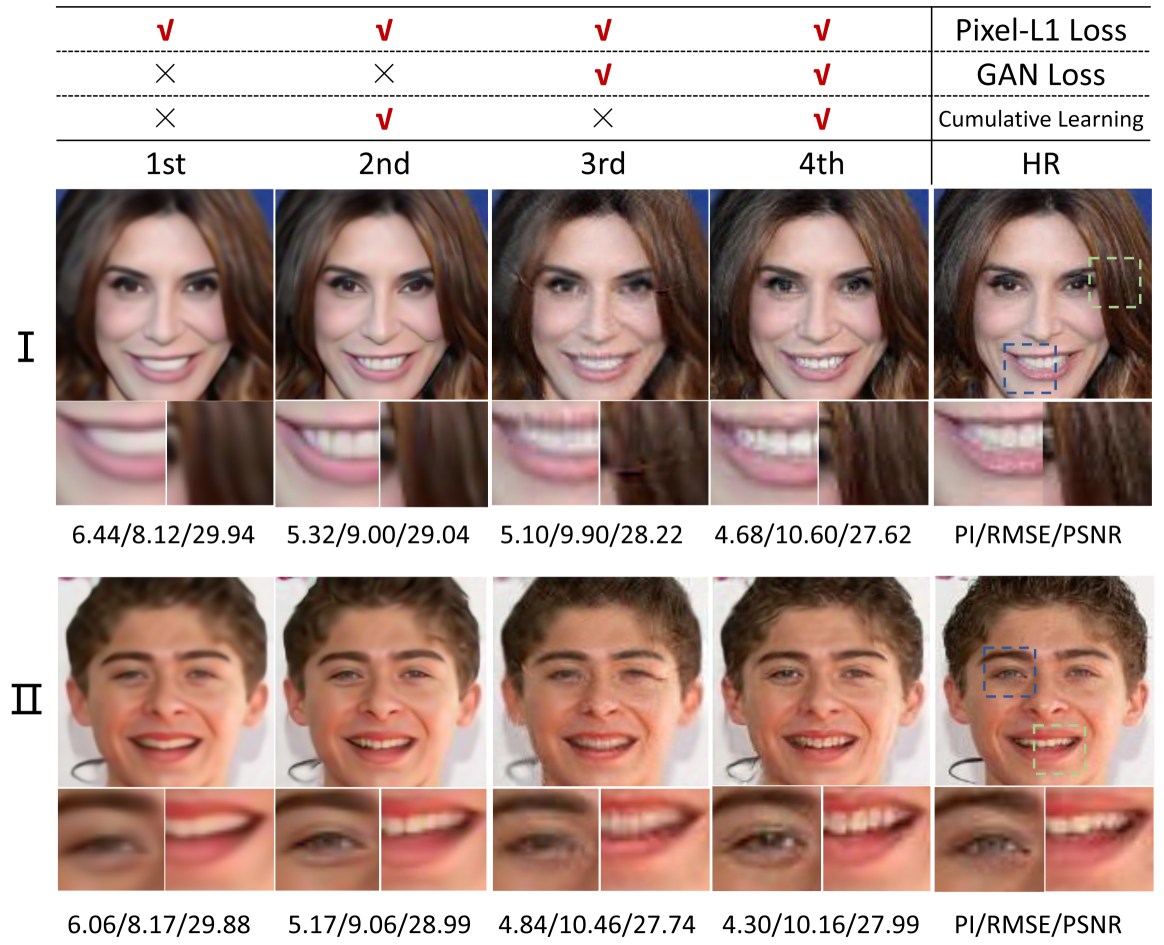}} %
	\caption{
		Overall comparisons for showing the effects of each part in
PCA-SRGAN. Each column represents a model with its configurations in the top.
	} %
	\label{Fig 1} %
\end{figure}

\begin{figure}[!h]
	%\centering
	\centerline{
		\includegraphics[width=0.50\textwidth,height=0.3\textwidth]{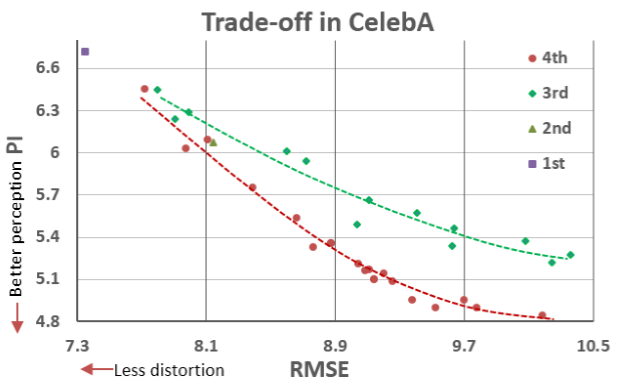}} %
	\caption{
			The comparison of quantitative Perception-Distortion trade-off. The GAN-based configurations can be plotted as the curves for the trade-off and other two configurations are plotted as single points in the figure. 
	} %
	\label{Fig 1} %
\end{figure}
The overall visual comparison is illustrated in Figure 6.
Only with the Pixel-L1 loss (see the 1st column) , the model generates smooth face without any texture. On the basis of the Pixel-L1 loss, clear edge is produced by utilizing the cumulative learning of PCA projection (see the 2nd column).  Combining the L1 loss with  GAN loss which is applied to the fixed 99.9\% PCA projection, the perceptual performance is improved but the GAN loss brings obvious noise and distortion (see the 3rd column). 
%At last, the proposed PCA-SRGAN containing all of three parts achieves better perceptual performance and generate realistic texture and details.
At last, under the same weights condition of GAN regularization, our model with cumulative learning obtains better visual performance and generates realistic texture (see the 4th column). 

Furthermore, a quantitative comparison of Perception-Distortion trade-off for GAN-based configurations has been also given to validate the performance of cumulative learning strategy. Obviously as shown in Figure 7, the model with cumulative discrimination (the red curve) overwhelms the same model without cumulative strategy (the green curve) by a much better trade-off boundary on whole regions.

In a word, the cumulative learning of PCA projection along with the adversarial discrimination in our model has suggested convincing effectiveness to enhance the face super resolution task.

%\section{The effect of Perception loss}

\section{Conclusion}
In this paper, a method named PCA-SRGAN using incremental orthogonal projection discrimination is proposed to enhance the performance of GAN on face SR task. 
We perform cumulative discrimination of orthogonal projections on PCA subspace to reduce the training difficulty and achieve precise and stable reconstruction without the help of perceptual (VGG) loss. 
Qualitative and quantitative comparisons have revealed the compelling performance of our model on super-resolving realistic face and the trade-off between perception and distortion. In the further research, we will develop our model by introducing other types of dictionary space or apply the model to other image processing tasks.

\vfill\pagebreak

\bibliographystyle{named}
\bibliography{ijcai20}

\end{document}